\documentclass[]{ceurart}

\usepackage{booktabs}
\usepackage{multirow}
\usepackage{pifont}
\usepackage{graphicx}
\usepackage{tabularx}
\usepackage[section]{placeins}
\usepackage{xurl}
\usepackage{needspace}
\newcommand{\cmark}{\ding{51}}
\newcommand{\rot}[1]{\rotatebox{90}{\scriptsize\bfseries #1}}
\newcommand{\syskw}[1]{\textit{(#1)}}
\newcommand{\team}[1]{\textit{#1}}
\newcommand{\system}[1]{\textit{#1}}
\newcommand{\entryspace}{\Needspace{9\baselineskip}\smallskip}

\begin{document}

\copyrightyear{2026}
\copyrightclause{Copyright for this paper by its authors.
  Use permitted under Creative Commons License Attribution 4.0
  International (CC BY 4.0).}
\conference{CLEF 2026 Working Notes, 21--24 September 2026, Jena, Germany}

\title{Overview of FinMMEval 2026 Task 2: Multilingual Financial Short-Answer Question Answering}

\author[1]{Zhuohan Xie}[email=zhuohan.xie@mbzuai.ac.ae]
\cormark[1]
\author[2]{Xueqing Peng}
\author[3]{Georgi Georgiev}
\author[3]{Dimitar Dimitrov}
\author[4]{Yuyang Dai}
\author[1]{Rania Elbadry}
\author[5]{Vanshikaa Jani}
\author[2]{Lingfei Qian}
\author[6]{Fan Zhang}
\author[2]{Jimin Huang}
\author[7]{Jiahui Geng}
\author[1,8]{Yankai Chen}
\author[8,9]{Ye Yuan}
\author[8,9]{Haolun Wu}
\author[4]{Yuxia Wang}
\author[3]{Ivan Koychev}
\author[1]{Veselin Stoyanov}
\author[10]{Mingzi Song}
\author[6]{Yu Chen}
\author[1,8,9]{Xue Liu}
\author[1]{Preslav Nakov}[email=Preslav.Nakov@mbzuai.ac.ae]

\address[1]{Mohamed bin Zayed University of Artificial Intelligence, Abu Dhabi, United Arab Emirates}
\address[2]{The Fin AI, United States}
\address[3]{FMI, Sofia University ``St. Kliment Ohridski'', Sofia, Bulgaria}
\address[4]{INSAIT, Sofia University ``St. Kliment Ohridski'', Sofia, Bulgaria}
\address[5]{University of Arizona, Tucson, United States}
\address[6]{The University of Tokyo, Tokyo, Japan}
\address[7]{Link{\"o}ping University, Link{\"o}ping, Sweden}
\address[8]{McGill University, Montreal, Canada}
\address[9]{Mila, Quebec AI Institute, Montreal, Canada}
\address[10]{Meiji Gakuin University, Tokyo, Japan}
\cortext[1]{Corresponding author.}

\begin{abstract}
FinMMEval 2026 Task 2 evaluates short-answer financial question answering over multilingual evidence.
Each final-test item pairs an English question with financial statements and news in English, Chinese, Japanese, Spanish, and Greek.
Participating systems submit one concise answer per item in JSONL format.
The final-test set contains 256 items, split evenly between easy and expert tiers; each tier contains four question templates instantiated over 32 company-report groups.
Gold answers were withheld during submission, and systems were ranked by macro-averaged item-level ROUGE-1 F1 against organizer-held reference answers.
The final leaderboard includes 12 ranked submissions.
The strongest systems are closely clustered, with the top four separated by less than one percentage point in ROUGE-1 F1.
The submitted system papers document retrieval-augmented generation, cross-lingual evidence handling, structured prompting, answer compression, and validation strategies.
\end{abstract}


\maketitle

\section{Introduction}

Financial question answering requires evidence selection, accounting and entity interpretation, and numerical or domain reasoning.
In a short-answer setting, systems must also compress the selected evidence without losing the financial facts required by the question.
Task 2 therefore evaluates concise answers by their lexical overlap with an organizer-held reference; this captures one dimension of answer quality rather than serving as a proxy for factual correctness~\cite{xie2023nextchapter}.
It is part of the broader FinMMEval lab at CLEF 2026~\cite{FinMMEval2026}, alongside companion tracks on multilingual multiple-choice QA in Task 1~\cite{FinMMEvalTask1Overview2026} and live trading agents in Task 3~\cite{FinMMEvalTask3Overview2026}.
Task 2 builds on the PolyFiQA resources from MultiFinBen~\cite{peng2025multifinben}, a benchmark suite for multilingual and multimodal financial applications.
Its hidden-test design requires systems to align financial statements with multilingual news and express the resulting evidence in an answer that can be evaluated against an organizer-held reference.

\section{Related Work}

Task 2 is related to financial question answering, multilingual financial NLP, and automatic short-answer evaluation.
FiQA introduced financial QA in an information-retrieval-oriented shared-task setting~\cite{maia2018fiqa}.
FinQA~\cite{chen2021finqa} and TAT-QA~\cite{zhu2021tatqa} focus on numerical and tabular reasoning over financial reports, while ConvFinQA adds a conversational setting in which systems must track evidence across turns~\cite{chen2022convfinqa}.
FinanceBench studies open-book financial QA over public-company information~\cite{islam2023financebench}.
FinChain~\cite{xie2025finchain} and RealFin~\cite{dai2026realfin} further emphasize verifiable reasoning steps and implicit financial assumptions.

Task 2 also relates to multilingual and document-grounded financial evaluation.
MultiFinBen extends financial evaluation to multilingual and multimodal applications~\cite{peng2025multifinben}.
SAHM provides a complementary Arabic financial reasoning benchmark~\cite{elbadry2026sahm}.
Finance-domain representation learning provides modeling context for financial reports, news, and terminology, from encoder-style FinBERT~\cite{liu2021finbert} to large-scale BloombergGPT~\cite{wu2023bloomberggpt}.
Instruction and evaluation resources such as PIXIU~\cite{xie2023pixiu} and FinGPT~\cite{yang2023fingpt} broaden this toolset for financial LLM systems.
Evidence selection is also central to intra-document evidence reranking for financial question answering in FinCARDS~\cite{zhou2026fincards} and to localized disclosure reporting in FinReporting~\cite{zhang2026finreporting}.
Herculean extends financial evaluation from static tasks to end-to-end agentic workflows~\cite{peng2026herculean}.
Where many prior settings focus on a fixed document collection or a single question-answering style, Task 2 emphasizes multilingual financial information needs and concise answer generation under a shared hidden-test protocol.

\section{Dataset}

\subsection{Data Collection}

The final-test set was constructed from organizer-held November versions of PolyFiQA-Easy and PolyFiQA-Expert.
PolyFiQA was introduced as part of MultiFinBen~\cite{peng2025multifinben}.
Each tier contained 204 candidate items.
We excluded 76 items whose source task identifier or full query matched the corresponding public PolyFiQA release, because those releases exposed both the questions and their reference answers.
The remaining 128 items per tier formed a 256-question hidden test.

The questions and answer instructions are in English.
The multilingual component lies in the evidence bundle: each item combines financial statements with financial news in English, Chinese, Japanese, Spanish, and Greek.
Easy questions ask for factual or numerical findings from the statements and supporting news evidence, whereas expert questions ask for news-based synthesis supported by financial-statement evidence.

\subsection{Data Annotations}

Each final-test item pairs a unique task identifier and difficulty tier with an English question and its multilingual evidence context.
Tier-specific answer templates distinguish easy responses, which provide an answer and news evidence, from expert responses, which provide an answer and supporting financial-statement evidence.
Reference answers were retained from the source PolyFiQA rows after overlap removal and remained organizer-held.
Each submission provided one non-empty answer per task identifier, with \texttt{None} reserved for cases in which no answer could be found.

\subsection{Data Statistics}

In each tier, 51 candidate source groups with four questions each produced 204 items; 19 groups already represented in the public release accounted for the 76 exclusions; and 32 retained groups yielded the remaining $204 - 76 = 128$ held-out items.
Easy and Expert use the same 32 retained groups, providing four questions from each tier per group and 256 final-test items overall (Table~\ref{tab:data}).

\begin{table}[!htbp]
\centering
\caption{Held-out test construction by tier. Public-overlap items matched the corresponding public PolyFiQA release and were excluded.}
\label{tab:data}
\small
\setlength{\tabcolsep}{4.5pt}
\begin{tabular}{@{}lrrrrr@{}}
\toprule
\textbf{Tier} & \makecell{\textbf{Candidate}\\\textbf{items}} & \makecell{\textbf{Public}\\\textbf{overlap}} & \makecell{\textbf{Held-out}\\\textbf{items}} & \makecell{\textbf{Retained}\\\textbf{groups}} & \makecell{\textbf{Items/}\\\textbf{group}} \\
\midrule
Easy & 204 & 76 & 128 & 32 & 4 \\
Expert & 204 & 76 & 128 & 32 & 4 \\
\midrule
Combined & 408 & 152 & 256 & 32 & 8 \\
\bottomrule
\end{tabular}
\end{table}

\section{Evaluation Framework}

\subsection{Task Organization}

Under the hidden-test protocol, each submission was a JSONL file containing exactly one non-empty answer for each of the 256 official task identifiers.
Reference answers and scores remained hidden until evaluation, and the final complete valid submission from each team entered the official leaderboard.

\subsection{Evaluation Measures}

The official metric is macro-averaged item-level ROUGE-1 F1~\cite{lin2004rouge}, computed between submitted answers and organizer-held reference answers.
The scorer processes the complete string stored in each \texttt{answer} field, including section labels and evidence text.
It applies Unicode NFKC normalization, lowercases the text, tokenizes words, numbers, CJK characters, Japanese kana, Arabic-script characters, and Greek/Latin ranges with a regular expression, and computes unigram multiset overlap.
For each item, unigram precision $P$ and recall $R$ are computed against the reference answer, and item-level F1 is:
\[
\mathrm{ROUGE\mbox{-}1\ F1} = \frac{2PR}{P+R}.
\]
Because all reference answers in the final test are non-empty, an item receives an F1 score of 0 when $P+R=0$.
The leaderboard score averages the 256 item-level ROUGE-1 F1 values; mean item-level precision and recall are reported as diagnostic components of unigram overlap.
The entire response contributes to ROUGE-1; the 100-word answer limit is a formatting guideline rather than a separate scoring term.

\subsection{Development References and Baselines}

Public PolyFiQA examples and the submission validator provided development references for answer formatting and end-to-end pipeline checks.
The final leaderboard reports participant systems only and does not include an organizer model baseline as a ranked entry.

\section{Results and Participant Approaches}

\subsection{Competition Results}

The final Task 2 ranking contains 12 complete submissions spanning direct extraction, retrieval, cross-lingual processing, structured prompting, and answer validation.
The top of the leaderboard is tightly clustered: \team{Calibrated Signals} ranked first with a macro-averaged ROUGE-1 F1 of 31.18\%, and only 0.79 percentage points separated the top four systems (Table~\ref{tab:results}).
The frozen leaderboard aggregates all 256 items; separate easy- and expert-tier scores were not released, so the official results do not support tier-specific performance comparisons.

\begin{table}[!htbp]
\centering
\caption{Official Task 2 final-test leaderboard with macro-averaged ROUGE-1 F1 and diagnostic precision and recall.}
\label{tab:results}
\small
\setlength{\tabcolsep}{6pt}
\begin{tabular}{@{}r l r r r@{}}
\toprule
\textbf{Rank} & \textbf{Team} & \textbf{F1 (\%)} & \textbf{Prec. (\%)} & \textbf{Rec. (\%)} \\
\midrule
1 & Calibrated Signals & 31.18 & 30.25 & 37.64 \\
2 & Pranshu Rastogi & 30.96 & 32.08 & 35.37 \\
3 & IGT & 30.71 & 28.21 & 40.44 \\
4 & AI\_TLfanClub & 30.39 & 36.47 & 30.70 \\
5 & pjmathematician & 29.72 & 27.81 & 37.98 \\
6 & DS@GT & 28.53 & 25.72 & 38.88 \\
7 & The Lab Rats & 27.34 & 27.37 & 34.17 \\
8 & Ethereum Team B CLEF & 25.46 & 30.86 & 26.42 \\
9 & TCLabs & 25.06 & 23.21 & 32.57 \\
10 & HU\_LLM\_Fin & 22.89 & 25.81 & 28.03 \\
11 & NLP-DE & 20.49 & 26.43 & 19.01 \\
12 & TextSentinels & 18.73 & 30.29 & 17.56 \\
\bottomrule
\end{tabular}
\end{table}
\FloatBarrier

\subsection{Analysis of Results}

The diagnostic precision and recall values reveal different lexical-overlap profiles.
\team{IGT} recorded the highest recall (40.44\%), whereas \team{AI\_TLfanClub} recorded the highest precision (36.47\%).
Because the official score is macro-averaged ROUGE-1 F1, different balances between unigram precision and recall can produce similar F1 values; neither diagnostic column directly measures factual correctness.
Because the release includes no organizer baseline, tier-specific results, or paired uncertainty analysis, the leaderboard supports comparisons among submitted systems but does not establish absolute task difficulty or statistically reliable differences among the tightly clustered leaders.

\subsection{Participant Approaches}

Table~\ref{tab:systems} summarizes components reported in ten Task 2 Working Notes papers.
Blank cells mean that a component was not reported.
Structured prompting appears throughout the documented systems, while evidence handling varies among direct extraction, retrieval, and reranking.

\begin{table}[!htbp]
\centering
\caption{Components reported by the ten documented Task 2 systems.}
\label{tab:systems}
\footnotesize
\setlength{\tabcolsep}{3.0pt}
\begin{tabular}{@{}>{\raggedright\arraybackslash}p{0.25\linewidth}|cccc|cc|ccc@{}}
\toprule
\multirow{2}{*}[-5.5ex]{\textbf{Team}} & \multicolumn{4}{c|}{\textbf{Evidence}} & \multicolumn{2}{c|}{\textbf{Reasoning}} & \multicolumn{3}{c}{\textbf{Output}} \\
\cmidrule(lr){2-5}\cmidrule(lr){6-7}\cmidrule(l){8-10}
& \rot{Extraction} & \rot{Retrieval} & \rot{Reranking} & \rot{Tools} & \rot{Agent graph} & \rot{Structured prompting} & \rot{Translation/routing} & \rot{Compression} & \rot{Validation/MBR} \\
\midrule
IGT~\cite{chiu2026igtTask2} & \cmark & & & & & \cmark & \cmark & & \\
Calibrated Signals~\cite{jain2026calibratedSignalsTask2} & & & & & & \cmark & & \cmark & \cmark \\
TextSentinels~\cite{durairaj2026textsentinelsTask2} & & \cmark & & & & \cmark & & & \\
AI\_TLfanClub~\cite{duc2026fiscoRag} & & \cmark & & & & \cmark & & & \cmark \\
DS@GT~\cite{liu2026finnexusTask2} & & \cmark & \cmark & \cmark & \cmark & \cmark & \cmark & & \cmark \\
The Lab Rats~\cite{gill2026labRatsTask2} & & & \cmark & & & \cmark & \cmark & \cmark & \\
Pranshu Rastogi~\cite{rastogi2026geminiFinMMEval} & \cmark & & & & & \cmark & \cmark & & \\
HU\_LLM\_Fin~\cite{affan2026huLLMFinTask2} & & \cmark & \cmark & & & \cmark & \cmark & & \\
TCLabs~\cite{pontes2026tclabsFinMMEval} & & & & & \cmark & \cmark & & & \\
pjmathematician~\cite{vachharajani2026pjmathematicianFinMMEval} & & & & & & \cmark & & \cmark & \cmark \\
\bottomrule
\end{tabular}
\end{table}
\FloatBarrier

\entryspace
\noindent\textbf{\team{IGT}~\cite{chiu2026igtTask2}}\\
\syskw{question-type prompting, targeted extraction, multilingual financial QA}\\
The system handles structured numerical questions with direct keyword and regular-expression extraction over filing text; for synthesis-oriented questions, it applies rule-based selection of multilingual passages before concise generation.
Its official implementation uses Claude Sonnet 4 through AWS Bedrock with type-specific prompts for the easy and expert question families.

\entryspace
\noindent\textbf{\team{Calibrated Signals}~\cite{jain2026calibratedSignalsTask2}}\\
\syskw{inference-only prompting, deterministic decoding, schema enforcement}\\
\team{Calibrated Signals} reports an inference-only system for cross-lingual financial question answering.
The system uses a fixed deployable prompt, deterministic decoding, schema enforcement, and one constrained retry when the required answer format is violated.

\entryspace
\noindent\textbf{\team{TextSentinels}~\cite{durairaj2026textsentinelsTask2}}\\
\syskw{BM25 retrieval, evidence preservation, multilingual QA}\\
\team{TextSentinels} builds a BM25-grounded method for multilingual financial QA.
Its final pipeline preserves chunk-local evidence, limits overly aggressive retrieval, and constrains generation with the selected passages.

\entryspace
\noindent\textbf{\team{AI\_TLfanClub}~\cite{duc2026fiscoRag}}\\
\syskw{FiSCO-RAG, cross-lingual evidence recall, BM25 retrieval, context packing}\\
\team{AI\_TLfanClub} proposes \system{FiSCO-RAG}, a retrieval-augmented approach designed for cross-lingual evidence recall in financial QA.
The system combines layout-aware chunking, CJK-aware BM25 retrieval, partition-aware context packing, schema-conditioned prompting, verifier filtering, and IDF-regularized minimum Bayes risk answer selection.

\entryspace
\noindent\textbf{\team{DS@GT}~\cite{liu2026finnexusTask2}}\\
\syskw{FinNexus, LangGraph orchestration, Weaviate retrieval, reranking, tool execution}\\
\team{DS@GT} introduces \system{FinNexus}, an agentic retrieval-augmented reranking system for multilingual financial QA.
The system uses LangGraph orchestration with Weaviate retrieval, reranking, and Python tool execution for numerical analysis.
Retrieved passages are evaluated by a judging agent before tool-assisted reasoning and short-answer generation.

\entryspace
\noindent\textbf{\team{The Lab Rats}~\cite{gill2026labRatsTask2}}\\
\syskw{GPT-4o-mini translation, evidence ranking, GPT-4o reasoning, answer compression}\\
\team{The Lab Rats} use GPT-4o-mini for translation and evidence ranking, GPT-4o for chain-of-thought reasoning, and GPT-4o-mini again for answer compression.
The system addresses cross-lingual inputs by normalizing the question context before reasoning and generation.
The final response is produced through a second call that compresses the reasoning output into the required short-answer format.

\entryspace
\noindent\textbf{\team{Pranshu Rastogi}~\cite{rastogi2026geminiFinMMEval}}\\
\syskw{Gemini, prompt engineering, tier-aware prompts, cross-lingual report/news QA}\\
\team{Pranshu Rastogi} uses prompt-engineered Gemini models for Task 2.
The system treats cross-lingual financial report and news QA as a prompt-design problem.
It uses separate prompts for easy and expert questions and targeted numerical-extraction passes where needed, without adding a retrieval component.

\entryspace
\noindent\textbf{\team{HU\_LLM\_Fin}~\cite{affan2026huLLMFinTask2}}\\
\syskw{hybrid RAG, row-aware chunking, partitioned reranking, mixture-of-experts}\\
\team{HU\_LLM\_Fin} reports a structure-aware hybrid RAG system using mixture-of-experts generation.
The system preserves table structure through row-aware chunking, combines dense and lexical retrieval, applies partitioned cross-encoder reranking, and generates answers with the mixture-of-experts model Llama 4 Scout using tier-specific prompts.

\entryspace
\noindent\textbf{\team{TCLabs}~\cite{pontes2026tclabsFinMMEval}}\\
\syskw{multilingual QA, expert agents, structured prompts, answer generation}\\
\team{TCLabs} describes a Task 2 multilingual QA system with separate news, report, and financial-reasoning experts before final answer generation.
An answer-generation agent consolidates the news and report analyses into a concise response.

\entryspace
\noindent\textbf{\team{pjmathematician}~\cite{vachharajani2026pjmathematicianFinMMEval}}\\
\syskw{Gemini, structured prompting, structural validation, answer compression}\\
The \team{pjmathematician} Task 2 system uses Gemini to produce concise, evidence-grounded answers from financial filings and multilingual news inputs.
Its validation step checks output structure and answer length, aiming to preserve the core evidence while satisfying the short-answer format.

Across the documented systems, recurring components included sparse or dense retrieval, evidence reranking, translation, tier-specific prompting, schema validation, and answer compression.
The tightly clustered leaders use prompt-only, extraction-oriented, and retrieval-augmented approaches, but results for complete systems do not isolate the components responsible for the small score differences.

\section{Conclusion and Future Work}

FinMMEval 2026 Task 2 evaluated systems on a 256-item hidden test of financial short-answer QA grounded in multilingual evidence and derived from PolyFiQA.
The official leaderboard, based on macro-averaged ROUGE-1 F1, contains 12 ranked submissions.
The documented systems include both retrieval-augmented and prompt-only approaches, with different balances between precision and recall under lexical-overlap scoring.
Task 2 is a compact hidden-test setting for open-ended financial QA beyond multiple-choice evaluation.
Future editions should report performance by question type, evidence language, and difficulty tier, and should complement ROUGE with semantic or human-audited factuality checks to better distinguish wording effects from financial understanding.

\Needspace{8\baselineskip}
\begin{acknowledgments}
We thank the CLEF 2026 organizers for hosting the lab and supporting the working-notes process.
We also thank all participating teams for submitting systems and providing feedback on the submission portals and evaluation workflow.
We are grateful to Georgi Georgiev for supporting the FinMMEval awards, which helped recognize strong participant submissions across the tasks.
The work of Dimitar Dimitrov and Ivan Koychev is partially supported by the project UNITe BG16RFPR002-1.014-0004 funded by PRIDST and also by the EU NextGenerationEU project, through the National Recovery and Resilience Plan of the Republic of Bulgaria, project SUMMIT, No.~BG-RRP-2.004-0008.
\end{acknowledgments}

\begin{aideclaration}
During the preparation of this work, the authors used OpenAI GPT-5.5 for grammar checks, wording revisions, style improvements, and LaTeX formatting assistance.
\end{aideclaration}

\bibliography{custom}

\end{document}